\documentclass[letterpaper, 10 pt, conference]{class/ieeeconf}  % Comment this line out
\IEEEoverridecommandlockouts                              % This command is only
\usepackage{xcolor}
\usepackage{siunitx}  
\usepackage{graphicx}
 \graphicspath{./figures/robot_demonstration}
\usepackage{amsmath}
\usepackage{amssymb}
\usepackage{latexsym}
\usepackage{url}
\usepackage{cite}
\usepackage{relsize}
\usepackage{multirow}
\usepackage{afterpage}
\usepackage{ifthen}
\usepackage{graphicx}
\usepackage{algpseudocode,algorithm,algorithmicx}
\usepackage{subfigure}
\usepackage{epstopdf}
\usepackage{stackengine}
\usepackage{textcomp} % new

\usepackage{gensymb}
\usepackage[bookmarks=false, linkcolor=blue, urlcolor=blue, citecolor=blue]{hyperref} 
\usepackage{booktabs}
\usepackage{makecell}
\usepackage{hhline}
\usepackage{courier}
\usepackage{lipsum}
\usepackage{bm}
\usepackage{pifont}% http://ctan.org/pkg/pifont
\newcommand{\cmark}{\ding{51}}%
\newcommand{\xmark}{\ding{55}}%
\usepackage{xcolor} % new

\hypersetup{
    colorlinks=true,
    linkcolor=red,
    filecolor=magenta,      
    urlcolor=blue,
    pdfstartview={FitH}    }

\graphicspath{{./figures/}}

\makeatletter
\renewcommand{\dddot}[1]{%
  {\mathop{\kern\z@#1}\limits^{\makebox[0pt][c]{\vbox to-1.4\ex@{\kern-\tw@\ex@
   \hbox{\normalfont ...}\vss}}}}}
\makeatother
\providecommand{\FullStop}{\text{~\@.\xspace}}
\providecommand{\Comma}{\text{~,\xspace}}
\newcommand{\vect}[1]{\boldsymbol{\mathbf{#1}}}
\newcommand{\dist}[1]{\left\lVert #1 \right\rVert}
\providecommand{\qdot}{\dot{\vect{q}}}

\usepackage{xspace}

\providecommand{\kuka}{\textsc{KUKA} LBR iiwa 14 R820\xspace}

    \title{\LARGE \bf Language-Driven Closed-Loop Grasping with Model-Predictive Trajectory Replanning}

\author{Huy Hoang Nguyen$^{1,\dag}$, Minh Nhat Vu$^{2,4,\dag,\ddag}$, Florian Beck$^{2,\dag}$, Gerald Ebmer$^2$, Anh Nguyen$^3$, Andreas Kugi$^{2,4}$
\thanks{$^1$ Faculty of Informatics, Eötvös Loránd University, Budapest, Hungary {\tt jsocgh@inf.elte.hu}}
\thanks{$^2$ Automation \& Control Institute (ACIN), TU Wien, Vienna, Austria {\tt \{vu,beck,ebmer,kugi\}@acin.tuwien.ac.at}}
\thanks{$^3$ Department of Computer Science, University of Liverpool, UK {\tt anh.nguyen@liverpool.ac.uk}}
\thanks{$^4$ Austrian Institute of Technology (AIT) GmbH, Vienna, Austria {\tt \{Minh.Vu,Andreas.Kugi\}@ait.ac.at}}
\thanks{$\dag$ Equal contribution, $\ddag$ Corresponding author}
}

\begin{document}
% Macros

\newtheorem{problem}{Problem}
\newtheorem{lemma}{Lemma}
\newtheorem{theorem}[lemma]{Theorem}
\newtheorem{claim}{Claim}
\newtheorem{corollary}[lemma]{Corollary}
\newtheorem{definition}[lemma]{Definition}
\newtheorem{proposition}[lemma]{Proposition}
\newtheorem{remark}[lemma]{Remark}
\newenvironment{LabeledProof}[1]{\noindent{\it Proof of #1: }}{\qed}

\def\beq#1\eeq{\begin{equation}#1\end{equation}}
\def\bea#1\eea{\begin{align}#1\end{align}}
\def\beg#1\eeg{\begin{gather}#1\end{gather}}
\def\beqs#1\eeqs{\begin{equation*}#1\end{equation*}}
\def\beas#1\eeas{\begin{align*}#1\end{align*}}
\def\begs#1\eegs{\begin{gather*}#1\end{gather*}}

\newcommand{\poly}{\mathrm{poly}}
\newcommand{\eps}{\epsilon}
\newcommand{\e}{\epsilon}
\newcommand{\polylog}{\mathrm{polylog}}
\newcommand{\rob}[1]{\left( #1 \right)} %Round Brackets
\newcommand{\sqb}[1]{\left[ #1 \right]} %square Brackets
\newcommand{\cub}[1]{\left\{ #1 \right\} } %curly brackets
\newcommand{\rb}[1]{\left( #1 \right)} %Round
\newcommand{\abs}[1]{\left| #1 \right|} %| |
\newcommand{\zo}{\{0, 1\}}
\newcommand{\zonzo}{\zo^n \to \zo}
\newcommand{\zokzo}{\zo^k \to \zo}
\newcommand{\zot}{\{0,1,2\}}
\newcommand{\en}[1]{\marginpar{\textbf{#1}}}
\newcommand{\efn}[1]{\footnote{\textbf{#1}}}
\newcommand{\vecbm}[1]{\boldmath{#1}} %more general (handles greek letters)
\newcommand{\uvec}[1]{\hat{\vec{#1}}}
\newcommand{\thv}{\vecbm{\theta}}
\newcommand{\junk}[1]{}
\newcommand{\var}{\mathop{\mathrm{var}}}
\newcommand{\rank}{\mathop{\mathrm{rank}}}
\newcommand{\diag}{\mathop{\mathrm{diag}}}
\newcommand{\tr}{\mathop{\mathrm{tr}}}
\newcommand{\acos}{\mathop{\mathrm{acos}}}
\newcommand{\atantwo}{\mathop{\mathrm{atan2}}}
\newcommand{\SVD}{\mathop{\mathrm{SVD}}}
\newcommand{\quadf}{\mathop{\mathrm{q}}}
\newcommand{\linterp}{\mathop{\mathrm{l}}}
\newcommand{\sgn}{\mathop{\mathrm{sign}}}
\newcommand{\sym}{\mathop{\mathrm{sym}}}
\newcommand{\avg}{\mathop{\mathrm{avg}}}
\newcommand{\mean}{\mathop{\mathrm{mean}}}
\newcommand{\erf}{\mathop{\mathrm{erf}}}
\newcommand{\grad}{\nabla}
\newcommand{\R}{\mathbb{R}}
\newcommand{\defeq}{\triangleq}
\newcommand{\dims}[2]{[#1\!\times\!#2]}
\newcommand{\sdims}[2]{\mathsmaller{#1\!\times\!#2}}
\newcommand{\udims}[3]{#1}
\newcommand{\udimst}[4]{#1}
\newcommand{\com}[1]{\rhd\text{\emph{#1}}}
\newcommand{\ind}{\hspace{1em}}
\newcommand{\argmin}[1]{\underset{#1}{\operatorname{argmin}}}
\newcommand{\floor}[1]{\left\lfloor{#1}\right\rfloor}
\newcommand{\step}[1]{\vspace{0.5em}\noindent{#1}}
\newcommand{\quat}[1]{\ensuremath{\mathring{\mathbf{#1}}}}
\newcommand{\norm}[1]{\left\lVert#1\right\rVert}
\newcommand{\ignore}[1]{}
\newcommand{\specialcell}[2][c]{\begin{tabular}[#1]{@{}c@{}}#2\end{tabular}}
\newcommand*\Let[2]{\State #1 $\gets$ #2}
\newcommand{\algorithmicbreak}{\textbf{break}}
\newcommand{\Break}{\State \algorithmicbreak}
\newcommand{\ra}[1]{\renewcommand{\arraystretch}{#1}}

\renewcommand{\vec}[1]{\mathbf{#1}} %looks better

\algdef{S}[FOR]{ForEach}[1]{\algorithmicforeach\ #1\ \algorithmicdo}
\algnewcommand\algorithmicforeach{\textbf{for each}}
\algrenewcommand\algorithmicrequire{\textbf{Require:}}
\algrenewcommand\algorithmicensure{\textbf{Ensure:}}
\algnewcommand\algorithmicinput{\textbf{Input:}}
\algnewcommand\INPUT{\item[\algorithmicinput]}
\algnewcommand\algorithmicoutput{\textbf{Output:}}
\algnewcommand\OUTPUT{\item[\algorithmicoutput]}

\maketitle
\thispagestyle{empty}
\pagestyle{empty}

%%%%%%%%%%%%%%%%%%%%%%%%%%%%%%%%%%%%%%%%%%%%%%%%%%%%%%%%%%%%%%%%%%%%%%%%%%%%%%%%
\begin{abstract}
Combining a vision module inside a closed-loop control system for the \emph{seamless movement} of a robot in a manipulation task is challenging due to the inconsistent update rates between utilized modules. This task is even more difficult in a dynamic environment, e.g., objects are moving. 
This paper presents a \emph{modular} zero-shot framework for language-driven manipulation of (dynamic) objects through a closed-loop control system with real-time trajectory replanning and an online 6D object pose localization. 
We segment an object within $\SI{0.5}{\second}$ by leveraging a vision language model via language commands. 
Then, guided by natural language commands, a closed-loop system, including a unified pose estimation and tracking and online trajectory planning, is utilized to continuously track this object and compute the optimal trajectory in real-time. Our proposed zero-shot framework provides a smooth trajectory that avoids jerky movements and ensures the robot can grasp a non-stationary object. Experimental results demonstrate the real-time capability of the proposed zero-shot modular framework to accurately and efficiently grasp moving objects. The framework achieves update rates of up to \SI{30}{\hertz} for the online 6D pose localization module and \SI{10}{\hertz} for the receding-horizon trajectory optimization. These advantages highlight the modular framework's potential applications in robotics and human-robot interaction; see the video at \href{https://www.acin.tuwien.ac.at/en/6e64/}{https://www.acin.tuwien.ac.at/en/6e64/}. 
\end{abstract}

%%%%%%%%%%%%%%%%%%%%%%%%%%%%%%%%%%%%%%%%%%%%%%%%%%%%%%%%%%%%%%%%%%%%%%%%%%%%%%%%

\section{INTRODUCTION} \label{Sec:Intro}
%DO NOT DELETE \note{figure link: shorturl.at/glALZ} 
Integrating artificial intelligence (AI) in robotic systems transforms human-robot collaboration, particularly in dynamic and complex environments. % such as healthcare and manufacturing. 
Traditional robotic systems typically rely on pre-programmed tasks and controlled environments with limited variability. However, real-world applications demand higher adaptability and precision, especially when robots need to interact with humans in real-time.
Imagine we want an assistant robot to grasp a hammer among a clutter of workshop objects such as scissors, scales, screwdrivers, or pliers. 
The robot must be capable of continuously detecting and grasping the hammer, even if the hammer can be moved.
%While humans can naturally understand how to grasp a hammer when instructed, it is still challenging for robots to identify precise grasping actions for objects based on natural language commands, a task known as language-driven grasp detection~\cite{shridhar2022cliport}. 
This simple example encompasses three challenges, i.e., \textit{natural language understanding}, \textit{real-time object localization and tracking}, and \textit{real-time trajectory planning}. Despite recent advancements, bridging the gap between language, vision, and control in real-world robotic experiments remains challenging. %~\cite{yang2023pave}. 
Recent works, e.g., LLaRP~\cite{szot2023large} and Open X-embodiment~\cite{vuong2023open}, often focus on an end-to-end approach relying on pre-defined large skill datasets. This reliance on individual skill acquisition is a significant bottleneck for application in various systems, primarily due to mismatches in dynamic models, which result in jerky motions of the robot. Furthermore, these end-to-end approaches often neglect kinematic and dynamic constraints, leading to unsafe and inaccurate robot motions. These requirements are significant concerns for meeting industrial standards, where precise and reliable operations are critical. 
Thus, in this work, we focus on creating a zero-shot modular framework utilizing a vast knowledge of foundation models for vision-language understanding and smoothly controlling the robot at a granular action level, bypassing the need for extensive data collection or manual programming of individual basic movements. Also, with the modular fashion, we can integrate these individual system constraints into the trajectory optimization process to ensure that the robot's movements are safe and precise.

\begin{figure*}[t]
\centering
 \includegraphics[width=\textwidth]{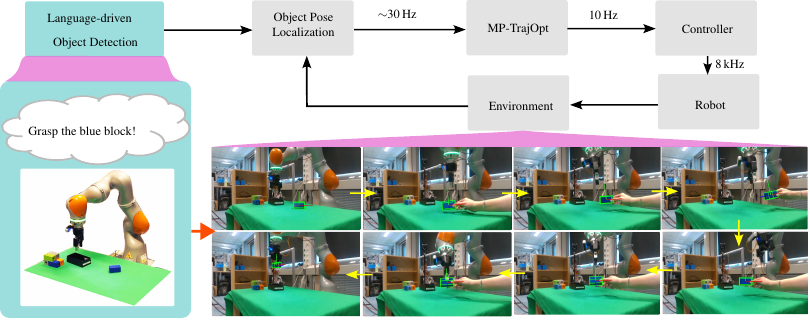}
    \vspace{1pt}
\caption{Overview of the proposed zero-shot framework: The top row shows the modules of the proposed framework, which run with different update rates, i.e., the language-driven object detection module (low frequency), the object poses localization module (medium frequency), the MP-TrajOpt module (medium frequency), and the controller (very high frequency). Note that grey blocks are a closed control loop. The overlay images in the second row illustrate an execution of the experimental evaluation.}
\label{fig: overview learning}
\end{figure*}

To address the challenge above in language-driven grasping tasks, we propose a modular zero-shot framework comprising \textit{(i)} a vision-language model to analyze the human prompts and to determine the mentioned object, \textit{(ii)} a vision module to localize and track the object in real-time, and \textit{(iii)} the receding-horizon trajectory optimization to plan a smooth trajectory toward the object. Specifically, we first utilize open-vocabulary detectors, such as %NanoOWL \cite{NanoOWL} 
and OWLv2 \cite{minderer2023scaling}, leveraging contrastively pre-trained image and text encoders on the large-scale image-text data set RegionCLIP \cite{zhong2022regionclip}. The open-vocabulary detector allows us to extract spatial-geometric information of relevant objects from natural language commands. Next, we employ the unified foundation model for 6D object pose localization, FoundationPose \cite{wen2023foundationpose}, to efficiently estimate and continuously track the object's 6D pose. Given that the object is not stationary, we employ our recent work, the model-predictive trajectory optimization (MP-TrajOpt)\cite{beck2024rec}, to dynamically guide the robot toward the object. Figure \ref{fig: overview learning} illustrates an overview of the proposed framework, described in detail in Section \ref{Sec:method}.

The structure of the paper is as follows: We begin with a discussion of the related work in the next section, followed by the presentation of our proposed framework in Section \ref{Sec:method}. This framework includes three modules: language-driven object detection, object pose localization, and model-predictive trajectory optimization. In Section \ref{Sec:exp}, we delve into the experiments and analysis, including comparisons with other approaches and the limitations of our framework. Finally, in Section \ref{Sec:con}, we conclude our work and provide some remarks.

\section{Related Work} \label{Sec:rw}
\textbf{Foundation Models for Robotic Manipulation.}
Various methods have been proposed to integrate foundation models into robotic manipulation~\cite{szot2023large, rana2023sayplan}. For instance, Ha et al.\cite{ha2023scaling} use large language models (LLMs) as knowledge bases to decide tasks and utilize sampling-based motion planners to create comprehensive manipulation skill datasets. These datasets can then be used to train diffusion policies for diverse manipulation tasks using natural language commands. While the trend of applying foundation models to robotics is growing \cite{jin2024robotgpt}, it still faces significant uncertainties~\cite{liu2023llm+}.
One major issue is the need for improved functional competence in these models~\cite{liu2023llm+}, which limits their ability to tackle novel planning problems that require an understanding of real-world dynamics~\cite{mahowald2023dissociating}. Additionally, enabling language models to perceive physical environments often involves providing textual descriptions and incorporating vision during decoding~\cite{driess2023palm}. To close the perception-action loop, an embodied language model must also be capable of action, typically achieved through a library of pre-defined primitives. However, these primitives vary across robotic platforms, often requiring a large dataset to generalize effectively. Unlike previous methods, we leverage pre-trained language models for their extensive open-world knowledge and focus on the complex task of planning for dynamic object manipulation. Our approach incorporates system dynamic constraints, enabling a seamless integration of perception and action. This method addresses the limitations of previous models by enhancing functional competence and facilitating real-time interaction and adaptation.

\textbf{Language-driven object detection.} %\textcolor{red}{TODO: correct this!}
Traditionally, language-driven object detection methods were designed for limited categories, a.k.a. closed-vocabulary object detection, like DETR \cite{zhu2020deformable}. Recently, large vision-language models (VLMs) like CLIP \cite{radford2021learning} have opened doors to open-vocabulary object detection. Unlike traditional methods, open-vocabulary object detection methods OWLv2 \cite{minderer2023scaling} and RegionCLIP \cite{zhong2022regionclip} are not restricted to pre-defined categories. Open-vocabulary object detection allows them to excel in tasks like zero-shot object detection and segmentation based on natural language descriptions. This shift towards open-vocabulary object detection \cite{minderer2023scaling} extends beyond identifying objects. It has applications in related areas like affordance detection \cite{nguyen2023open}, where the focus is on identifying potential actions. Inspired by OWLv2, we utilize this method in our framework to enable zero-shot object detection via natural language commands. 

\textbf{Object pose localization.}
%\lipsum[1]
Recent advancements in object pose estimation have focused on developing techniques that handle novel objects without requiring retraining, addressing a key challenge in applying deep learning models to industrial settings. Nguyen et al. \cite{nguyen2023gigapose} introduced GigaPose, simplifying the pose estimation process to a single correspondence match between a query image and a set of templates. Recently, Wen et al. \cite{wen2023foundationpose} presented FoundationPose, a unified 6D object pose estimation and tracking model, utilizing large-scale synthetic training data and a novel transformer-based architecture. FoundationPose supports both model-based and model-free setups with novel objects. However, manual segmentation of objects in the first frame from pre-recorded videos is still required. In this work, we integrate a language-driven object detection module that could enable end-to-end applications from human commands to robotic grasping actions in a closed-loop system.

\textbf{Real-time trajectory optimization.} 
%\lipsum[1]
In recent years, significant interest has been in extending trajectory optimization to online planning using a receding horizon. This interest spans gradient-based methods \cite{Schoels2020a} and sampling-based approaches \cite{Bhardwaj2022}. 
However, point-to-point planning is often not enough for manipulation tasks.
Additional constraints, such as pre-grasp points, must be considered too.
The sequence-of-constraints Model Predictive Control (MPC) introduced by Toussaint et al. \cite{Toussaint2022} addresses task and motion planning (TAMP) in three phases: initially, task planning generates waypoints; subsequently, the timing of these waypoints is optimized to form a reference trajectory; finally, MPC tracks this reference trajectory to compute collision-free paths over a short planning horizon. Note that this approach requires a global reference to incorporate waypoints into the MPC. Tackling this issue, in our recent work \cite{beck2024rec}, we propose the model-predictive trajectory optimization MP-TrajOpt that avoids calculating a reference trajectory for waypoint timing. Instead, this MPC approach employs a cost-to-go analysis for waypoints. The advantages of integrating MP-TrajOpt are twofold. First, to complete a general grasping task, the robot must navigate specific waypoints to avoid collisions between its gripper and the object. For example, the robot should pass through a waypoint \SI{0.05}{\meter} above the object along the $z$-axis in a top-down grasp, as illustrated in Figure \ref{fig: pre-grasp}. Second, passing waypoints with non-zero velocity allows for a more natural and smooth grasping action, which differs from most state-of-the-art methods \cite{tang2023graspgpt,huang2023voxposer}.

%In this work, 
Leveraging the vast knowledge from foundation models, we tackle the language-driven tasks in this world. Our contributions are summarized as follows: 
\begin{itemize}
    \item
    We present a zero-shot modular framework for end-to-end language-driven closed-loop grasping tasks that operate without training data. It is important to note that the primary goal of this work is to offer a comprehensive implementation rather than serve as an evaluation benchmark.
\item
   We implement the online version of object pose localization, specifically FoundationPose \cite{wen2023foundationpose}, which includes synchronizing RGB and depth sensors data and automating the segmentation process via the language-driven object detection module, OWLv2 \cite{minderer2023scaling}. 
\item 
   We provide extensive demonstrations with real-world objects and comparisons with state-of-the-art methods \cite{huang2023voxposer}. Additionally, we open-source our implementation at \href{https://language-driven-closed-loop-grasping.github.io/}{language-driven-closed-loop-grasping.github.io} to support community development. 
\end{itemize} 
\section{Methods} \label{Sec:method}
%\textbf{Problem description.}
This section explains the three main modules: language-driven object detection, real-time object pose localization, and online trajectory optimization (MP-TrajOpt), illustrated in Figure \ref{fig: overview learning}. Consider a scenario in a workshop where a human instructs a robot to grasp a tool with a known CAD model and place it in a pre-defined location. First, the language-driven object detection module processes the image to determine the 2D location of the tool and generate a binary mask. This mask, combined with the CAD model, is used by the real-time object pose localization module to estimate the initial pose of the tool. A Kalman filter refines the object's pose for smoothness. Finally, the online trajectory optimization module plans and executes the grasp and placement actions, ensuring smooth and collision-free movements. 
\begin{figure*}[t]
    \centering
    \includegraphics[width=\textwidth]{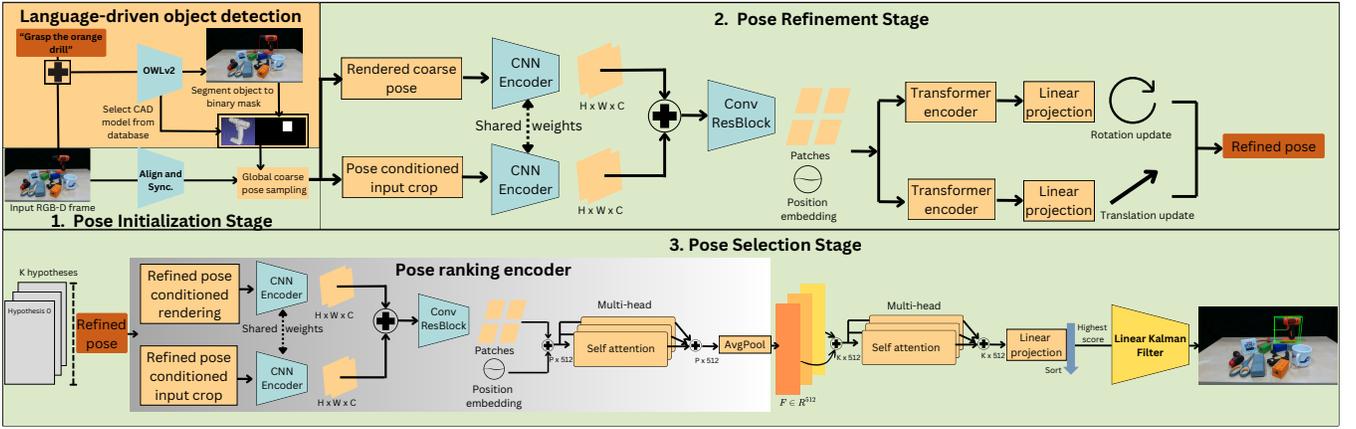}
    \vspace{1pt}
    \caption{
    Flow chart of the language-driven object detection and object pose localization module: The vision module OWLv2 \cite{minderer2023scaling} detects the object's 2D location to create a binary mask, then uses this mask with a CAD model for initial pose estimation \cite{wen2023foundationpose}, refined by a Kalman filter. The light orange block indicates the language-driven object detection module, while the light green block presents the three object pose localization stages. 
    }
    \label{fig: flow chart}
\end{figure*}

\subsection{Language-Driven Object Detection}
To determine the 2D location of a target object specified in a natural language command, such as "Grasp the orange drill," we first extract key phrases from the command, in this case, "orange drill." Concurrently, an RGB image is input into the OWLv2 model \cite{minderer2023scaling} to generate pseudo-box annotations for image-text pairs, as illustrated in Figure \ref{fig: flow chart}.
Although we utilize the pre-trained weights of the OWLv2 model without additional training, this approach is efficient and effective at pinpointing the object's center within the image. Compared to more complex segmentation methods, such as SAM \cite{kirillov2023segment} or CNOS \cite{nguyen2023cnos}, which require several seconds for inference, the OWLv2 model can ground the object within \SI{0.5}{\second} across all experimental cases.
A binary mask is generated upon obtaining the target object's bounding box. This binary mask serves as the input for the subsequent object pose localization process.
\subsection{Object Pose Localization Module}

Inspired by the state-of-the-art pose estimation method by Wen et al.\cite{wen2023foundationpose}, we present a real-time object pose estimation and tracking framework employing a three-stage pipeline, as depicted in the schematic of Figure \ref{fig: flow chart}.

In the pose initialization stage, we align and synchronize data. %from the depth sensor and the RGB sensor of an RGB-D camera. %, thereby striking a delicate balance between precision and computational efficiency. 
Subsequently, leveraging the object detection mask generated by the method mentioned earlier, we formulate an initial coarse pose estimation. This initial estimation is the input to a pose refinement network in the subsequent stage.
The architecture of this refinement network, depicted in the second stage of Figure \ref{fig: flow chart}, is designed to infer translation updates $\Delta \mathbf{p} \in \mathbb{R}^3$ and rotation updates $\Delta \mathbf{R} \in SO(3)$ of the corresponding object. Each update is individually processed through a transformer encoder\cite{vaswani2023attention} and then linearly projected to the output dimension. Both $\Delta \mathbf{p}$ and $\Delta \mathbf{R}$ are expressed relative to the camera frame. The initial coarse pose $[\mathbf{R} | \mathbf{p}] \in SE(3)$ is refined utilizing the following update equations~\cite{wen2023foundationpose}
\begin{align}
\mathbf{p}^+ &= \mathbf{p} + \Delta \mathbf{p} \\
\mathbf{R}^+ &= \Delta \mathbf{R} \oplus \mathbf{R},
\end{align}
where $\oplus$ indicates the update operation on $SO(3)$. This disentangled representation approach effectively decouples the translation update from the orientation update. The refinement network is trained under supervision~\cite{wen2023foundationpose} using the L2 loss function:
\begin{equation}
\mathcal{L}_{\text{refine}} = w_1 |\Delta \mathbf{p} - \Delta \overline{\mathbf{p}}|^2 + w_2 |\Delta \mathbf{R} - \Delta \overline{\mathbf{R}}|^2,
\end{equation}
where $\overline{\mathbf{p}}$ and $\overline{\mathbf{R}}$ denote the ground truth values, and $w_1$ and $w_2$ are empirically determined weights used to balance the loss terms, both set to $1$.

Following the refinement of pose hypotheses, a hierarchical pose ranking network computes their respective scores in the pose selection stage of Figure \ref{fig: flow chart}. By employing the hierarchical comparison and contrast validation approach proposed in\cite{wen2023foundationpose}, the network computes the loss for each pair of hypotheses, ultimately selecting the final pose estimate with the highest score. Notably, this network employs the same backbone architecture for feature extraction as the refinement network but with slight modifications by adding the multi-head self-attention. The extracted features are concatenated, tokenized, and processed through a multi-head self-attention mechanism and a pose ranking encoder to generate a feature embedding $\mathcal{F} \in \mathbb{R}^{512}$ across all pose hypotheses. Then, multi-head self-attention is applied to the concatenated feature embedding $\mathrm{F} = \begin{bmatrix} \mathcal{F}_0 & \dots & \mathcal{F}_{K-1} \end{bmatrix}^\top \in \mathbb{R}^{K \times 512}$, capturing pose alignment information from all hypotheses. Finally, the attended features are linearly projected to produce the scores $\mathrm{S} \in \mathbb{R}^{K}$, which are associated with the pose hypotheses. By selecting the highest scores for the pose hypotheses, we obtain accurate and meaningful pose estimations, ensuring the reliability of the final pose output.

Next, we employ a Kalman filter (KF) \cite{kalman1960linear} to smoothen the object's pose estimation. Triple integrator dynamics are used as a process model with white Gaussian noise as input to describe the object's position, $\mathbf{p}$, and orientation $\mathbf{R}$, parameterized by the vector of Euler angles $\mathbf{o} = [\psi,\phi,\theta]^\mathrm{T}$. Thus, the state vector for the process model reads as
\begin{equation}
    \bm{\xi}^\mathrm{T} = [\mathbf{p}^\mathrm{T},\dot{\mathbf{p}}^\mathrm{T},\ddot{\mathbf{p}}^\mathrm{T},\mathbf{o}^\mathrm{T},\dot{\mathbf{o}}^\mathrm{T},\ddot{\mathbf{o}}^\mathrm{T}]   \:,
\end{equation}
%where $\mathbf{o} = [\psi,\phi,\theta]^\mathrm{T}$ is the vector of three Euler angles computed from the object's orientation $\mathbf{R}$ \cite{siciliano2008springer}. 
%The process model for the KF with the sampling time $\Delta t$, under the constant acceleration assumption, at time index $k$ is
and the zero-order-hold discrete-time state-space formulation yields 
\begin{equation}
    \bm{\xi}_k = \mathbf{A} \bm{\xi}_{k-1} + \mathbf{w}_{k-1}\:,
\end{equation}
%with $\mathbf{A} = [1,\:\Delta t,\:0.5\Delta t^2,\:1,\:\Delta t,\:0.5\Delta t^2]^\mathrm{T}\otimes \mathbf{I}_3$, 
with the sampling time $\Delta t$, the time index k, and
\begin{equation}
    \mathbf{A} = \begin{bmatrix}
        1 & \Delta t & \frac{\Delta t^2}{2} \\
        0 & 1 & \Delta t \\
        0 & 0 & 1
    \end{bmatrix} \otimes \mathbf{I}_3 \Comma
\end{equation}
where $\otimes$ is the Kronecker product, and $\mathbf{I}_3$ is the identity matrix of size $3$.
The vector $\mathbf{w}$ is white Gaussian noise with the positive definite covariance matrix $\mathbf{Q}_w$. 
The measurement model is
\begin{equation}
    [\mathbf{p}_k^\mathrm{T},\mathbf{o}_k^\mathrm{T}] = \mathbf{H}\bm{\xi}_k +\mathbf{v}_k\:,
\end{equation}
with 
$\mathbf{H} = [1,\:0,\:0,\:1,\:0,\:0]^\mathrm{T}\otimes \mathbf{I}_3$, and $\mathbf{v}$ is white Gaussian measurement noise with the positive definite covariance matrix $\mathbf{Q}_v$. %We consider the following steps
The Kalman Filter equations for the estimated state $\hat{\bm{\xi}}_k$ read as
\begin{subequations}
\begin{align}
    \hat{\bm{\xi}}_k^- &= \mathbf{A}\hat{\bm{\xi}}_{k-1}\:, \\ \mathbf{P}_k^{-} &= \mathbf{A}\mathbf{P}_{k-1}\mathbf{A}^\mathrm{T} + \mathbf{Q}_w\:, \\
    \mathbf{K}_k &= \mathbf{P}_k^{-}\mathbf{H}^\mathrm{T}(\mathbf{H}\mathbf{P}_k^{-}\mathbf{H}^\mathrm{T}+\mathbf{Q}_v)^{-1} \:,\\
    \hat{\bm{\xi}}_k &= \hat{\bm{\xi}}_k^- + \mathbf{K}_k([\mathbf{p}_k^\mathrm{T},\mathbf{o}_k^\mathrm{T}]^\mathrm{T} - \mathbf{H}\hat{\bm{\xi}}_k^-) \:,\\
    \mathbf{P}_k &= (\mathbf{I} - \mathbf{K}_k\mathbf{H})\mathbf{P}_k^- \:\: .
\end{align}
\end{subequations}
%to update $\hat{\bm{\xi}}_k$ while eliminating jerky measurements from the pose selection stage. 
%Note that $\mathbf{Q}_w$ and $\mathbf{R}_m$ are the user-defined process and measurement noise covariance, respectively. 
%. Configured with an 18-state model encompassing position, velocity, and acceleration for both translation and rotation, the LKF is initialized with appropriate transition matrices, measurement matrices, process noise covariances, measurement noise covariances, and error covariances. 
%Subsequently, the pose estimation is propagated through the Kalman Filter, yielding a smoothed estimate of the pose by predicting the next state.

%\textbf{The vision part, Kalman part, and MP-TrajOpt are not linked together.}

\subsection{Model Predictive Trajectory Optimization (MP-TrajOpt)}

\subsubsection{Mathematical Model}
\label{sec: robot model}

The rigid-body dynamical model of a fully actuated robot is given by 
\begin{align}
  \label{eqn:rigid_body_dyn}
 \vect{M}(\vect{q})\ddot{\vect{q}} + \vect{C}(\vect{q}, \dot{\vect{q}})\dot{\vect{q}} + \vect{g}(\vect{q}) = \vect{\tau}\Comma
\end{align}
with the generalized coordinates $\vect{q} \in \mathbb{R}^m$, the positive definite mass matrix $\vect{M}(\vect{q})$, the matrix of Coriolis and centrifugal terms $\vect{C}(\vect{q}, \qdot)$, the gravitational forces $\vect{g}(\vect{q})$, and the generalized torques $\vect{\tau}$. 
%We consider only a rigid-body model for simplicity of presentation, neglecting elastic joints. 
%For the full elastic joint model, the reader is referred to~\cite{Ott2008} for a detailed derivation of the model and the controller.

The Cartesian end-effector pose of the robot, obtained by the forward kinematics of the robot, is described as a homogeneous transformation and given by
\begin{align}
  \vect{T}_\mathrm{e}(\vect{q}) = \begin{bmatrix}
                                \vect{R}_\mathrm{e}(\vect{q}) & \vect{p}_\mathrm{e}(\vect{q}) \\
                                \vect{0} & 1
                             \end{bmatrix}\Comma
\end{align}
containing the position $\vect{p}_\mathrm{e} \in \mathbb{R}^3$ and the rotation matrix $\vect{R}_\mathrm{e}(\vect{q}) \in \mathrm{SO}(3)$, and $\vect{0}$ denotes a zero matrix of matching size. 

The MP-TrajOpt from \cite{beck2024rec} requires a goal and waypoints in the joint space. 
We obtain those joint space solutions from the object pose localization using an analytic inverse kinematics solution of the \kuka \cite{vu2023machine}.
Compared to numerical methods, the computational efficiency of the analytical solution is much better.
Fast computation time is crucial for real-time applications.

\subsubsection{Control}
\label{sec:control}
We compensate the nonlinear robot dynamics with the computed torque state-feedback law
\begin{align}
  \label{eqn:computed_torque}
  \vect{\tau} = \vect{M}(\vect{q})\vect{u} + \vect{C}(\vect{q}, \dot{\vect{q}})\dot{\vect{q}} + \vect{g}(\vect{q})\Comma
\end{align}
which simplifies the dynamic model~(\ref{eqn:rigid_body_dyn}) to the double integrator system
\begin{align}
  \label{eqn:double_integrator}
  \ddot{\vect{q}} = \vect{u}\Comma
\end{align}
with the new control input $\vect{u}$. 
The remaining linear dynamics~(\ref{eqn:double_integrator}) are stabilized using
\begin{align}
  \label{eqn:control_law}
  \vect{u} &= \ddot{\vect{q}}_{\mathrm{d}} - \vect{K}_\mathrm{v} (\dot{\vect{q}} - \dot{\vect{q}}_{\mathrm{d}}) - \vect{K}_\mathrm{d} (\vect{q} - \vect{q}_{\mathrm{d}}) \Comma
\end{align}
where $\vect{K}_\mathrm{v}$ and $\vect{K}_\mathrm{d}$ are suitable gain matrices, making the closed-loop trajectory error dynamics exponentially stable. The desired reference trajectory in the joint space is given by $\vect{q}_{\mathrm{d}}(t)$ with its time derivatives $\dot{\vect{q}}_{\mathrm{d}}(t)$ and $\ddot{\vect{q}}_{\mathrm{d}}(t)$ and the time $t$.

\subsubsection{Planning}
\label{sec:model_planning}

The controller~\eqref{eqn:control_law} requires at least two times continuously differentiable reference trajectory from the planner. For additional smoothness, we use triple integrator dynamics for the planning algorithm. Thus, the state vector is defined as $\vect{x}^\mathrm{T} = [\vect{q}^\mathrm{T}, \dot{\vect{q}}^\mathrm{T}, \ddot{\vect{q}}^\mathrm{T}]$ with the input $\vect{u} = \dddot{\vect{q}}$.
Assuming piecewise-linear inputs $\vect{u}_k$ with the sampling time $h$ leads to the first-order-hold discrete-time state-space formulation
\begin{align}
  \vect{x}_{k + 1} = \vect{\Phi} \vect{x}_k + \vect{\Gamma}_1 \vect{u}_k  + \vect{\Gamma}_2 \vect{u}_{k + 1}\label{eqn:sys_d} \Comma
\end{align}
where
\begin{align}
  \vect{\Phi} &= \begin{bmatrix}
                  1 & h & \frac{h^2}{2} \\
                  0 & 1 & h \\
                  0 & 0 & 1    
                \end{bmatrix}\otimes \vect{I}_{m}\Comma\quad\nonumber \\
  \vect{\Gamma}_1 &= \begin{bmatrix}
                  \frac{h^3}{8} \\
                  \frac{h^2}{3} \\
                  \frac{h}{2}
               \end{bmatrix} \otimes \vect{I}_{m} \Comma\quad
  \vect{\Gamma}_2 = \begin{bmatrix}
                  \frac{h^3}{24} \\
                  \frac{h^2}{6} \\
                  \frac{h}{2}
               \end{bmatrix} \otimes \vect{I}_{m}\FullStop\label{eqn:sys_d_mat}
\end{align} 
The symbol $\otimes$ denotes the Kronecker product, and $\vect{I}_m$ is the identity matrix of size $m$.

We summarize the model-predictive trajectory planning (MP-TrajOpt) approach from our recent work~\cite{beck2024rec}.
We calculate the inverse kinematics solution~\cite{vu2023machine} of the pre-grasp and grasp pose and the pre-place and place pose to obtain a joint space waypoint and goal configuration $\vect{q}_\mathrm{w}$ and $\vect{q}_\mathrm{g}$.
To consider the waypoint along the way to the goal, we split the planning horizon of $N$ sampling points into two parts, namely $0, \dots, N_\mathrm{s} - 1$ and $N_\mathrm{s}, \dots, N - 1$ as soon as the waypoint $\vect{q}_\mathrm{w}$ is reachable within the horizon.
If the waypoint is reachable, the planning algorithm introduces a constraint for the waypoint to pass it with a certain tolerance.
Similarly, the planner introduces an endpoint constraint for the goal $\vect{q}_\mathrm{g}$ when it is reachable within the horizon.

The optimal system state $\vect{x}_{0 | n}, \dots, \vect{x}_{N - 1 | n}$ and input $\vect{u}_{0|n}, \dots, \vect{u}_{N - 1 | n}$ trajectory for the $n$-th iteration of the planner are obtained from the discrete-time optimization problem
\begin{subequations}
  \label{eqn:jerk_opt}
  \begin{alignat}{2}
    &\min_{\substack{\vect{x}_{0 | n}, \dots, \vect{x}_{N - 1 | n}, \\\vect{u}_{0 | n}, \dots, \vect{u}_{N - 1 | n}}} &&\sum_{k = 0}^{N_{\mathrm{s}} - 1}  w_1 l_1(\vect{x}_{k|n}) + \sum_{k = N_{\mathrm{s}}}^{N - 1}  w_2 l_2(\vect{x}_{k|n}) \nonumber \\ &\quad && + \sum_{k = 0}^{N - 1} \dist{\vect{u}_{k|n}}_2^2 + w_3 l_{\mathrm{col}}(\vect{x}_{k|n}) \label{eqn:opt_cost}\\
    &\quad \text{s.t.} && \vect{x}_{k + 1 | n} = \vect{\Phi} \vect{x}_{k | n } + \vect{\Gamma}_1 \vect{u}_{k | n} + \vect{\Gamma}_2 \vect{u}_{k + 1 | n},\nonumber \\ &\quad && k = 0, \dots, N - 2 \label{eqn:opt_dyn}\\
    & \quad && \vect{x}_{0 | n} = \vect{x}_{1 | n - 1},\quad \vect{u}_{0 | n} = \vect{u}_{1 | n - 1} \label{eqn:init_cond_1}\\
    & \quad && \vect{x}_{N - 1 | n} = \vect{\Phi} \vect{x}_{N - 1 | n},\quad \vect{u}_{N - 1 | n} = \vect{0}\label{eqn:steady_state_x}\\
    &                  \quad && \underline{\vect{x}} \le \vect{x}_{k | n} \le \overline{\vect{x}},\quad \underline{\vect{u}} \le \vect{u}_{k | n} \le \overline{\vect{u}} \label{eqn:x_limit} \\
    & \quad && \vect{q}_{N_{\mathrm{s}} - 1} \in \mathcal{Q}_{\mathrm{w}},\quad \vect{q}_{N - 1} \in \mathcal{Q}_{\mathrm{g}} \label{eqn:end_point_way} 
  \end{alignat}
\end{subequations}
where the cost terms $l_1$ and $l_2$ drive the trajectory to the waypoint and the goal, respectively.
The regularization cost on $\vect{u}_{k|n}$ ensures smooth trajectories, and $l_{\mathrm{col}}$ is a collision avoidance term.
The weights $w_1, w_2, w_3 > 0$ are tuning parameters.
The trajectory must adhere to the system dynamics \eqref{eqn:opt_dyn} with the initial states obtained by the previous planner iteration \eqref{eqn:init_cond_1}.
The endpoint is constrained to an equilibrium point in \eqref{eqn:steady_state_x}, and the states and inputs are bounded using upper $\overline{\vect{x}}$, $\overline{\vect{u}}$, and lower $\underline{\vect{x}}$, $\underline{\vect{u}}$ bounds.
The endpoints of the split horizon are constrained to a waypoint tolerance set $\mathcal{Q}_\mathrm{w}$ and a goal tolerance set $\mathcal{Q}_\mathrm{g}$, respectively.
For details on computing the horizon lengths $N_{\mathrm{s}}$ and $N$, refer to~\cite{beck2024rec}.

\begin{figure}[t]
\centering
\includegraphics[width=0.48\textwidth]{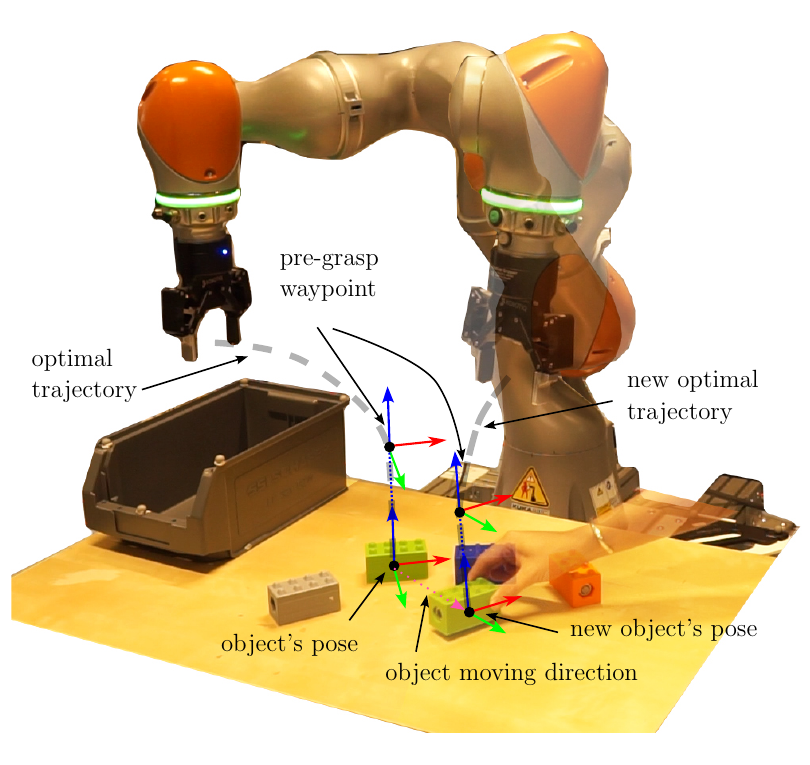}
\caption{Illustration of the pre-grasp waypoint located above the object's pose: The $x$-, $y$-, and $z$-axes are colored in red, green, and blue, respectively.}
\label{fig: pre-grasp}
\end{figure}

%\subsection{Task Planning \textcolor{red}{name?}}
We use a state machine to guide the robot through the pick-and-place task. 
First, the robot must approach the object to grasp it.
We derive a grasp pose and a pre-grasp waypoint from the object pose.
The pre-grasp pose is placed above the object to ensure a proper approach direction and a collision-free grasp. Figure \ref{fig: pre-grasp} illustrates an example of this pre-grasp waypoint. 
Note that moving the object further than a tolerance dynamically changes these poses.
When the robot end-effector reaches the grasp pose, the robot grasps the object and can plan toward the place location.
Similarly to the grasp poses, a pre-place and place pose are defined to place the object.

\section{Experiments} \label{Sec:exp}
\begin{figure}[h]
    \centering
    \includegraphics[width=0.8\linewidth]{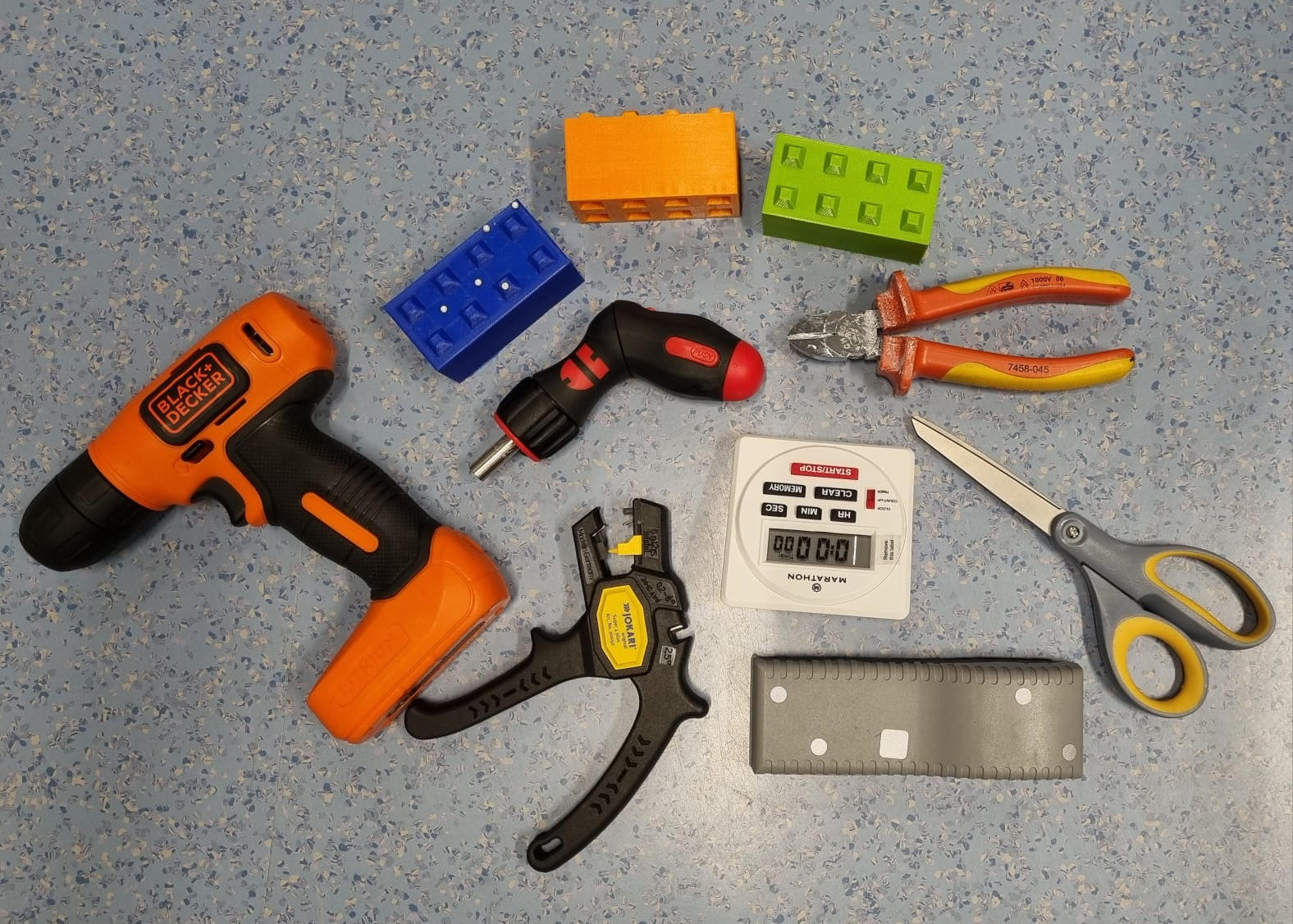}
    \vspace*{2mm}
    \caption{Set of ten objects used in the robotic experiment.}
    \label{fig: 10 objects}
\end{figure}

%The proposed framework is tested 
%An example of experiment setups 
We experimentally evaluate the proposed framework for grasping tasks in real-world scenarios with a 7-DoF \kuka and ten novel objects, captured in Figure \ref{fig: 10 objects}. We utilize the D435i RealSense camera in an eye-to-hand setup where this camera observes the robot within its workspace, as shown in Figure \ref{fig: pre-grasp}. 
Three modules of the proposed framework, see Figure \ref{fig: overview learning}, are executed on a desktop PC with an Intel Core i7 10700K CPU and 32 GB of RAM. Inference processes of the language-driven module and the object pose localization run on the NVIDIA RTX 3080 GPU with 8GB of VRAM. 
The optimization problem described in Equation~(\ref{eqn:jerk_opt}) is implemented as a Python-based ROS node utilizing CasADi~\cite{Andersson2019}. It is solved using the nonlinear interior point solver IPOPT~\cite{Waechter2006} and the MA57 linear solver, running on an Ubuntu 22.04 system. This implementation achieves planning times under $\SI{100}{\milli\second}$, which includes the online solution of the analytic inverse kinematics for new Cartesian waypoints and a desired goal~\cite{vu2023machine}. The solution nearest to the previous waypoint is selected in the inverse kinematics using a least-squares criterion for consistency.

\subsection{Statistical Results of Vision Modules}
Considering the test setup depicted in Figure \ref{fig: pre-grasp} and all ten objects shown in Figure \ref{fig: 10 objects}, we conducted statistical analysis on the inference speeds and success rates of using another state-of-the-art language-driven vision module, namely CNOS \cite{nguyen2023cnos} with FastSAM \cite{kirillov2023segment}, in conjunction with the presented object localization module. For all test cases, the prompt "Please grasp the [object]" was used, with the placeholder [object] being replaced by terms such as "orange drill," "blue block," "orange block," "green block," etc. Each object underwent five trials. Table \ref{table: stat-vision-modules} presents the computation times of the inference processes of the vision modules using different GPUs. The term "oFP \& KF" refers to the online version of FoundationPose with a Kalman filter, which is utilized in our object pose localization module. The CNOS and FastSAM models are pretty slow, requiring several seconds to segment objects through language commands. Conversely, the lightweight OWLv2 requires approximately $\SI{0.57}{\second}$ to compute a binary mask of the segmented object via a prompt. With an Nvidia RTX 4090, an update rate of approximately $\SI{5}{\hertz}$ can be achieved using the language-driven module employing OWLv2 \cite{minderer2023scaling} and approximately $\SI{30}{\hertz}$ with the proposed object localization module. Under identical lighting conditions and prompts across 50 trials, OWLv2 detected objects in 46 instances, whereas the other methods succeeded in only 40 cases. It is important to note that once an object is detected and the binary mask is provided, the object localization model can reliably track the object.
\begin{table}[h]
    \caption{\label{table: stat-vision-modules} Statistical results on inference speed of different modules.}
    \vspace{2ex}
    \centering
    \renewcommand\tabcolsep{4pt}
    \hspace{1ex}
    \begin{tabular}{@{}rccc@{}}
        \toprule
        Hardware
          & 
          \begin{tabular}{@{}c@{}}CNOS \cite{nguyen2023cnos} \&  \\ FastSAM \cite{kirillov2023segment}\end{tabular}  &  OWLv2\cite{minderer2023scaling} & \begin{tabular}{@{}c@{}}oFP \cite{wen2023foundationpose}  \\ \& KF\end{tabular} \\
        \midrule
        Nvidia RTX 3080  & \SI{5.02}{\second}  & \SI{0.57}{\second} &  \SI{0.083}{\second} \\
        Nvidia RTX 4080S & \SI{3.75}{\second}  & \SI{0.24}{\second} & \SI{0.04}{\second}\\ 
        Nvidia RTX 4090 & \SI{2.81}{\second}  & \SI{0.18}{\second} & \SI{0.033}{\second}\\
        Success rate & 0.8  & 0.92 & 1\\
        %Success rate &0.27 & 0.30 \\
    \bottomrule
    \end{tabular}
\vspace{-2ex}
\end{table}
\subsection{Qualitative Results of the Performance of the Object Localization Module}
\begin{figure}[t]
\centering
\includegraphics[width=0.48\textwidth]{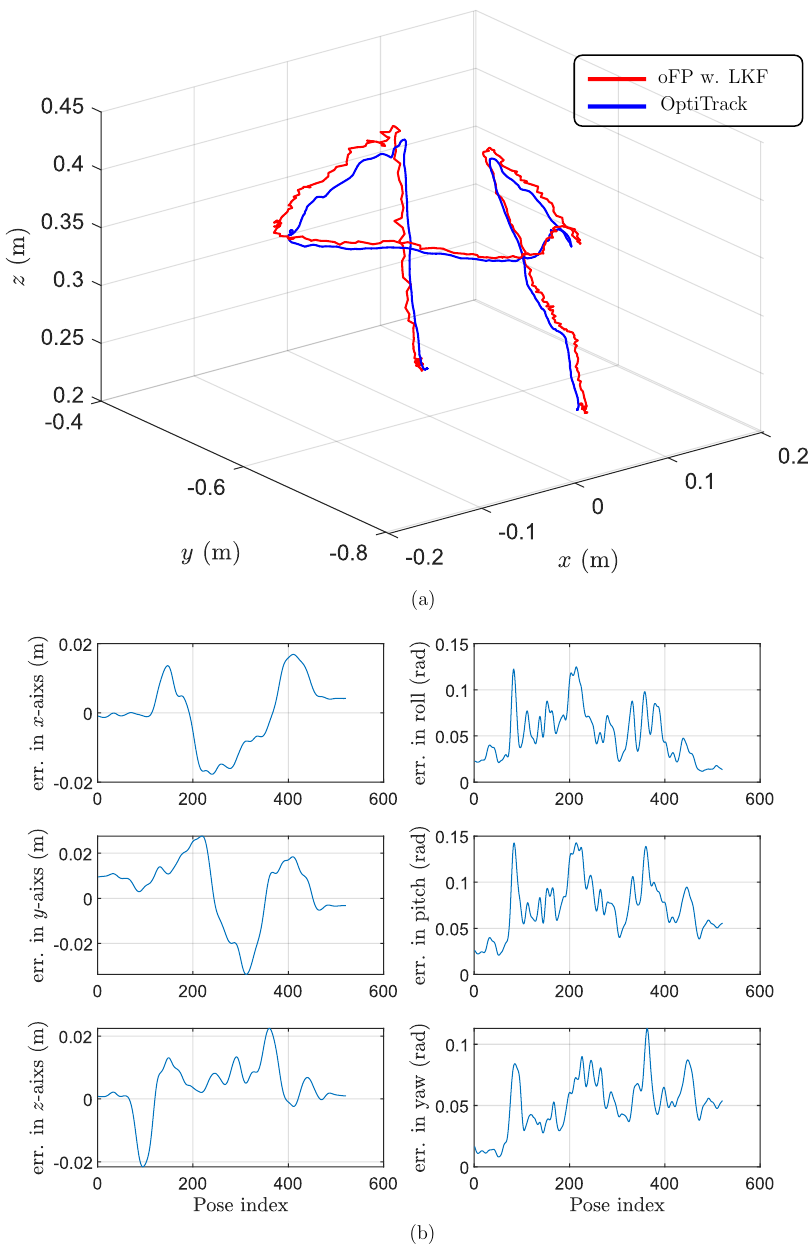}
    \vspace{1pt}
\caption{(a) 3D trajectory plot. (b) Errors plot}
\label{fig: accuracy}
\end{figure}

The qualitative performance of the object localization module was evaluated using ground truth data from an OptiTrack system. This assessment aimed to verify the accuracy of the object localization module. Figure \ref{fig: accuracy} presents the tracking results for the blue Lego block (depicted in Figure \ref{fig: 10 objects}). The 3D position trajectory of this Lego block closely matches the measurements obtained from the OptiTrack system. These measurements are transformed to the camera origin, approximately \SI{1}{\meter} from the object. The maximum positional error observed is approximately \SI{0.02}{\meter}, representing 2\% relative to the distance to the camera. The orientation error is constrained, with a maximum error of \SI{0.15}{\radian} in the roll angle. This evaluation is influenced by the jerky movements of the human hand while holding the object and the intense IR light emitted by the OptiTrack system. The errors are significantly lower when the object remains stationary, as illustrated in Figure \ref{fig: accuracy}. Notably, the vision module demonstrates promising accuracy in dynamic scenarios and can meet industrial standards in static scenarios.

\subsection{Robotic Experiments}
We evaluated ten novel objects, depicted in Figure \ref{fig: 10 objects}, using our proposed modular framework, AnyGrasp \cite{fang2023anygrasp}, LGD \cite{vuong2024lgd}, and VoxPoser \cite{huang2023voxposer}. Each object was subjected to five trials. It is important to note that only our framework supports moving objects; the other three methods were tested in static scenarios. Statistical results are presented in Table \ref{table: stat-result-modules}

AnyGrasp \cite{fang2023anygrasp}, which does not utilize natural language prompts, achieved 37 successful trials out of 50 without prompts. The LGD \cite{vuong2024lgd} approach encountered 27 failures due to incorrect object detections in cluttered scenarios. Unlike LGD \cite{vuong2024lgd} and AnyGrasp \cite{fang2023anygrasp}, both VoxPoser \cite{huang2023voxposer} and our modular framework consider the pre-grasp pose for all cases, as shown in Figure \ref{fig: pre-grasp}. Once the grasping pose is determined using AnyGrasp and LGD, the robot utilizes a standard inverse kinematic algorithm \cite{vu2023machine} alongside a joint space controller to guide its movements toward the object.

VoxPoser utilizes the standard task space controller \cite{ott2008cartesian}, which directs the robot toward the segmented object in the scenario. 
However, due to segmentation errors and occasional singularities in the task space controller, VoxPoser only achieved 33 successful cases. In contrast, our approach encountered only eight failures, primarily due to segmentation errors from OWLv2. Additionally, our proposed framework demonstrated the capability to replan in all 50 trials within dynamic scenarios. 

\begin{figure*}[t]
\centering
\includegraphics[width=\textwidth]{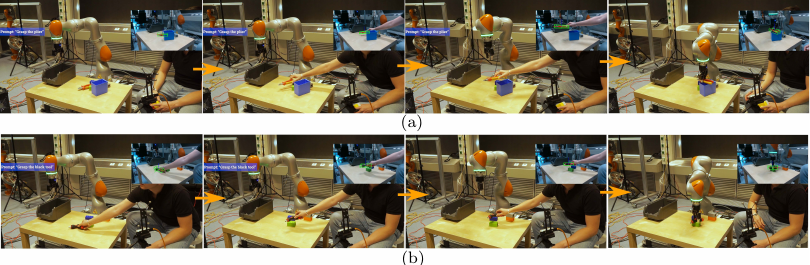}
\vspace{2ex}
\caption{Overlay images from demonstrations: (a) Prompt: "Grasp the plier" (b) Prompt: "Grasp the black tool"}. 
\label{fig: experiment prompt}
\end{figure*}

Figure \ref{fig: experiment prompt} shows overlay images from two experiments with our proposed framework. Small windows on the upper right at each image indicate the view from the Realsense camera. 
In Figure \ref{fig: experiment prompt} (a), with the given text prompt: ``Grasp the plier'', although the plier is occluded in the scenario, the proposed framework still outputs the correct pose of the plier, see the small window on the upper right of Figure \ref{fig: experiment prompt} (a). The proposed framework generally exhibits good results in localizing, tracking, and accurately grasping the object in a closed loop. 
More demonstrations are shown at  \href{https://www.acin.tuwien.ac.at/en/6e64/}{https://www.acin.tuwien.ac.at/en/6e64/}. 

\begin{table}[ht]
    \centering
    \caption{\label{table: stat-result-modules} Statistical results of experiments.}
    \vspace{2ex}
    \renewcommand
\tabcolsep{4pt}
\hspace{1ex}
    \begin{tabular}{@{}rcc@{}}
\toprule
Hardware
  & 
  Success rate  &  \begin{tabular}{@{}c@{}}Closed-loop \cite{wen2023foundationpose}  \\ replanning\end{tabular} \\
\midrule
AnyGrasp \cite{fang2023anygrasp} & 0.74  & \xmark \\
LGD \cite{vuong2024lgd} & 0.46  & \xmark \\
VoxPoser \cite{huang2023voxposer} & 0.66  & \xmark \\
Ours & 0.92  & \cmark \\
\bottomrule
\end{tabular}
\vspace{-2ex}
\end{table}

\subsection{Limitations}
Although we achieve good results, typical failure cases arise from a combination of challenges, including texture-less surfaces, severe occlusion, and limited edge cues, which lead to incorrect object detection and subsequent failed object localization. Additionally, the inference speed of the object localization process becomes a bottleneck when using low-profile GPUs, such as the NVIDIA RTX 3080. While upgrading to a more powerful GPU is an option, it is not always feasible, particularly in low-budget setups. Alternative solutions include incorporating prediction modules, such as those proposed by \cite{yan2022probabilistic}, or developing compact models for the object localization module. Furthermore, the proposed zero-shot framework cannot re-identify the object when object localization fails due to occlusions or rapid movement. Therefore, an upgraded framework with re-identification capabilities must address these issues. 

\section{Conclusions}\label{Sec:con}
We have introduced a novel zero-shot modular framework designed to address the challenges of language-driven manipulation of dynamic objects within a closed-loop control system. Our framework leverages a vision-language model for natural language processing and real-time object segmentation, allowing for the seamless integration of 6D object pose localization and receding-horizon trajectory optimization to achieve precise and smooth robotic movements. The intensive experimental results highlight the robustness and efficiency of our framework, achieving a 92\% success rate and superior inference speeds compared to state-of-the-art methods. Furthermore, the object localization module demonstrated high reliability with minimal positional and orientation errors, particularly in dynamic environments requiring real-time replanning. Future work will focus on further improving 6D pose estimation by continuously localizing object segmentation to prevent tracking loss and detecting affordance-based grasp points for objects through a language-driven approach.

%\section*{Acknowledgment}
%\addcontentsline{toc}{section}{Acknowledgment}
%\lipsum[1]

\bibliographystyle{class/IEEEtran}
\bibliography{class/IEEEabrv,class/reference}
   
\end{document}